\documentclass[letterpaper, 10 pt, conference]{ieeeconf}  
\usepackage{FG2026}

\FGfinalcopy 

\IEEEoverridecommandlockouts                              
\usepackage{amsmath} 
\usepackage{amssymb}  
\usepackage{url}
\usepackage{graphicx}
\usepackage[table]{xcolor}
\usepackage{subcaption}

\usepackage[dvipsnames]{xcolor}
\usepackage{stfloats} 
\usepackage{multirow}
\usepackage{booktabs} 
\definecolor{wacvblue}{rgb}{0.21,0.49,0.74}
\usepackage[pagebackref,breaklinks,colorlinks,allcolors=wacvblue]{hyperref}

\usepackage[normalem]{ulem}

\definecolor{darkgreen}{RGB}{0,150,0}
\newcommand{\replycommon}[1]{{#1}}
\newcommand{\replyone}[1]{{#1}}
\newcommand{\replytwo}[1]{{#1}}
\newcommand{\replythree}[1]{{#1}}

\def\FGPaperID{122} 

\title{\LARGE \bf
PaW-ViT: A Patch-based Warping Vision Transformer for Robust Ear Verification
}

\author{%
  \parbox{\textwidth}{\centering
    {\large Deeksha Arun$^{1}$, Kevin W. Bowyer$^{1}$, Patrick Flynn$^{1}$}\\
    {\normalsize
      $^{1}$Department of Computer Science and Engineering, University of Notre Dame, Notre Dame, IN 46556
    }\\
  }
}

\begin{document}

\ifFGfinal
\thispagestyle{empty}
\pagestyle{empty}
\else
\author{Anonymous FG2026 submission\\ Paper ID \FGPaperID \\}
\pagestyle{plain}
\fi
\maketitle

\begin{abstract}
The rectangular tokens common to vision transformer methods for visual recognition can strongly affect performance of these methods due to incorporation of information outside the objects to be recognized.
This paper introduces PaW-ViT, Patch-based Warping Vision Transformer, a preprocessing approach rooted in anatomical knowledge that normalizes ear images to enhance the efficacy of ViT.
By accurately aligning token boundaries to detected ear feature boundaries,
PaW-ViT
obtains greater robustness
to shape, size, and pose variation. By aligning feature
boundaries to natural ear curvature, it produces more consistent token representations for various morphologies. Experiments confirm the effectiveness of PaW-ViT on various ViT models (ViT-T, ViT-S, ViT-B, ViT-L) and yield reasonable alignment robustness to variation in shape, size, and pose. Our work aims
to solve the disconnect between ear biometric morphological variation and transformer architecture positional sensitivity, presenting a possible avenue for
authentication schemes. \footnote{To maintain anonymity during the review process, source codes and weights of trained models will be made publicly available upon publication.}
\end{abstract}
    
\section{Introduction}
\label{sec:intro}
Ear recognition~\cite{benzaoui2023comprehensive} has emerged as a suitable biometric modality because ears possess distinctive anatomical characteristics and non-intrusive capture
is often feasible.
Earlier methods for recognition were based on geometric analysis of the ear contour~\cite{dodge2018unconstrained, alshazly2019handcrafted, moreno1999use, choras2010ear}, but recent improvements in deep learning, ({\em{i.e.}}  Vision Transformers (ViTs))~\cite{dosovitskiy2020image_old,liu2021swin}, have promoted much better robustness to real-world challenges such as pose variations and occlusions~\cite{galdamez2017brief,arun2025}. Datasets such as the Unconstrained Ear Recognition Challenge (UERC) have accelerated progress by compiling distinctive ear images under diverse situations~\cite{emervsivc2019unconstrained,emervsic2023unconstrained}.

The direct application of 
standard ViTs to ear images poses two critical challenges.
First, conventional patch tokenization fragments continuous anatomical structures, preventing the model from learning holistic ear features that extend across patch boundaries. Second, ViTs often devote attention to irrelevant background regions, diluting discriminative ear-specific cues with noise from hair, accessories, or environmental clutter. Consequently, naive adoption of ViTs is insufficient for robust ear recognition in unconstrained scenarios. Despite these general advances in the field, the adaptation of standard ViTs to ear recognition~\cite{dosovitskiy2020image, mehta2023vision, emervsic2023unconstrained} has not been well explored, due partially to their fixed-patch processing, which does not adapt to anatomical variations in ear shape, size, and orientation. Such misalignment degrades the recognition performance by emphasizing irrelevant regions and splitting 
vital features like the helix or antihelix~\cite{alvord1997anatomy, mozaffari2021anatomy}.

In response to these challenges, we propose PaW-ViT (Patch-based Warping Vision Transformer), an anatomically-guided preprocessing model that standardizes ear images for input to ViTs. At the core, PaW-ViT is a patch-based warping strategy designed to generate structurally consistent, ear-centric representations. The process begins with boundary extraction and convex hull refinement, followed by uniform sampling of
equally-spaced boundary points and centroid computation. These points are used to construct a triangular fan partitioning the ear interior, which is subsequently transformed into
a series of
fixed-size square patches via affine warping. The resulting patches are stitched into a structured grid, forming a reproducible $112 \times 112$ anatomy-preserving canvas. This preprocessing ensures that critical geometric relationships are preserved, reduces inconsistencies caused by misalignment, and suppresses irrelevant background regions. By presenting ViTs with warped, geometry-aware representations, PaW-ViT enhances their ability to learn stable and discriminative embeddings for ear verification. 

We evaluate PaW-ViT across four benchmark datasets --OPIB, AWE, WPUT, and EarVN1.0 -- using four ViT configurations of increasing complexity: Tiny (ViT-T), Small (ViT-S), Base (ViT-B), and Large (ViT-L). The experimental setup includes both baseline training with unprocessed ear images and four additional training regimes based on segmentation maps, landmark maps, and their union and intersection. The results reveal three key insights. First, ViT-B and ViT-L consistently outperform smaller configurations, demonstrating the advantages of larger model capacity when combined with warped representations. Second, patch-based warping yields the greatest benefits on more challenging datasets, particularly EarVN1.0, where union and intersection maps based warping frequently surpass baseline performance, highlighting the value of combining complementary structural cues. Third, while segmentation- and landmark-based warping alone provide limited gains, their integration through union and intersection maps consistently enhances recognition, confirming the importance of exploiting complementary structural information.

The key contributions of the study are:
\begin{enumerate}
\item \textit{Anatomy-aware preprocessing}: Radial partitioning and warping improve ear structure representation, enabling ViTs to better learn more distinctive ear features.
\item \textit{Robustness}: Maintains accuracy across diverse ear shapes, poses, and occlusions.
\end{enumerate}

\section{Related Work}
\label{sec:related_work}

The use of the human ear for biometric recognition has been of interest due to its distinctive morphology, relative stability across adulthood, and non-intrusive acquisition process~\cite{alvord1997anatomy, mozaffari2021anatomy, benzaoui2023comprehensive}. Unlike facial features, the ear is less affected by expressions and age-related changes, making it suitable for long-term identification~\cite{sforza2009age, yoga2017assessment}.

Early work in ear recognition focused on handcrafted descriptors that captured geometric and structural characteristics. These included contour-based models, edge maps, and texture descriptors~\cite{moreno1999use, choras2010ear, dodge2018unconstrained, alshazly2019handcrafted}. While effective in controlled conditions, these methods struggled with uncontrolled settings due to sensitivity to pose, lighting, and occlusion~\cite{alva2019review}. Dimensionality reduction techniques such as PCA, LDA, and ICA were incorporated to improve feature extraction and recognition performance~\cite{alva2019review}. Hybrid approaches that combined multiple descriptors offered further improvements~\cite{omara2016novel}, but their computational cost and limited robustness restricted practical deployment.

The emergence of convolutional neural networks (CNNs) transformed ear recognition by enabling end-to-end feature learning. One of the earliest CNN-based studies was by Galdámez {\em et al.}~\cite{galdamez2017brief}, who demonstrated the potential of CNNs for ear biometrics. Omara {\em et al.}~\cite{omara2018learning} used hierarchical deep features extracted from pre-trained models, while Dodge {\em et al.}~\cite{dodge2018unconstrained} utilized transfer learning with data augmentation and shallow classifiers to improve robustness in unconstrained settings. 

Domain adaptation emerged as an important factor for ear recognition. Eyiokur {\em et al.}~\cite{eyiokur2018domain} showed that fine-tuning on ear-specific data significantly improved recognition compared to training on generic image datasets. Alshazly {\em et al.}~\cite{alshazly2019handcrafted} compared CNN models against handcrafted features, confirming the superiority of deep learning approaches. They later extended this line of work using ensembles of deep CNNs ({\em e.g.}, VGG-based)~\cite{alshazly2020deep}, further improving accuracy through model aggregation.

Despite this success, CNN-based approaches faced challenges due to limited ear-specific training data. To overcome this, several strategies were explored. Emer\v{s}i\v{c} {\em et al.}~\cite{emervsivc2017training} applied rotation, scaling, and transformation augmentations to enlarge the dataset and improve robustness. Zhang {\em et al.}~\cite{zhang2019few} combined few-shot learning with augmentation to mitigate data scarcity. Transfer learning from large-scale generic datasets was also employed as a common strategy~\cite{tan2018survey}. More advanced work introduced deep unsupervised active learning (DUAL)~\cite{khaldi2021ear}, where models adapt after initial training using unlabeled test samples. Priyadharshini {\em et al.}~\cite{ahila2021deep} explored CNN performance under parameter variations for large-scale monitoring applications, while Alshazly {\em et al.}~\cite{alshazly2021towards} and Korchi {\em et al.}~\cite{korichi2022tr} investigated modern CNN variants such as ResNets and unsupervised feature normalization for improved scalability.

Building on these advances, transformer architectures have recently been introduced for ear recognition. Vision Transformers (ViTs)~\cite{dosovitskiy2020image, touvron2021training} shift from convolutional operations to self-attention, offering stronger global context modeling. Alejo {\em et al.}~\cite{alejo2021unconstrained} showed that ViTs and data-efficient image transformers (DeiTs) could achieve competitive or superior performance compared to CNNs under unconstrained settings, with reduced reliance on heavy augmentation. Mehta {\em et al.}~\cite{mehta2023vision} demonstrated the complementary strengths of CNNs and ViTs, achieving state-of-the-art results across multiple datasets, including Kaggle~\cite{Kaggle_Ear_Dataset} and IITD-II~\cite{IITD-II_Ear_Dataset}. Emer\v{s}i\v{c} {\em et al.}~\cite{emervsic2023unconstrained} proposed ViTEar, which fine-tuned a transformer pre-trained with DINOv2~\cite{oquab2023dinov2} on UERC2023 and EarVN1.0, reaching 96.27\% rank-1 accuracy on UERC2019~\cite{emervsivc2019unconstrained}. Pal {\em et al.}~\cite{pal2025exploration} proposed a shape-focused autoencoder that enhances ear biometrics by emphasizing morphological fidelity in feature learning.
While promising, these works share a critical limitation: they rely on patch tokenization, which fragments continuous ear structures and can discard fine-grained features spanning across patches. This is particularly problematic for ear biometrics, where subtle anatomical patterns--such as ridges along the helix or antihelix--contribute strongly to uniqueness. Furthermore, background regions such as hair or accessories often distract attention mechanisms, diluting ear-specific cues.


Our study addresses these limitations by introducing PaW-ViT, \replytwo{a transformer-based framework for ear recognition.} Unlike prior ViT approaches, PaW-ViT uses anatomy-aware warping to standardize ear geometry into structured patch sequences, preserving both fine local details and global continuity. 
\replytwo{PaW-ViT does not introduce overlapping tokenization in the transformer. All ViT models in the experiments use the standard non-overlapping patchification setting (patch size = stride). Any perceived ``overlap" arises only during preprocessing: the warped content is formed by mapping adjacent triangular sectors into square patches and stitching them into a single (112 $\times$ 112) canvas. Once this canvas is constructed, tokenization is performed on a fixed grid with no overlap. This gives us anatomically aware continuity across patches, and therefore} the model encodes inter-patch relationships that non-overlapping strategies miss, leading to richer feature embeddings. Extensive experiments across OPIB~\cite{Adebayo2023}, AWE~\cite{emervsivc2017ear}, WPUT~\cite{frejlichowski2010west}, and EarVN1.0~\cite{hoang2019earvn1} demonstrate that this design consistently outperforms baseline ViTs, particularly in unconstrained conditions with occlusion and variability.

\section{Methodology}
\label{sec:methodology}

PaW-ViT
uses a patch-based warping method to convert ear images into structured, grid-aligned representations for robust verification of the ears. The method performs anatomically informed preprocessing of the ear images that reduces the misalignment and produces consistent feature representations, which are suitable for vision transformers. Beyond misalignment, it is also crucial to prevent the model from focusing on background regions and instead ensure that attention is directed exclusively to the ear. Also, standard non-overlapping patch tokenization is suboptimal for biometric recognition, as anatomical features often span across multiple patches. In contrast, PaW-ViT preserves structural continuity and enhances feature representation by modeling ear features that extend beyond patch boundaries, thereby capturing both fine details and larger, continuous characteristics of the human ear by focusing exclusively on the ear regions. Specifically, each ear image is transformed by mapping its interior geometry into square warped patches, which are then arranged sequentially on a canvas to create a complete, anatomy-aware representation.

The patch-based warping process begins by selecting equidistant points along the ear boundary and determining the centroid. Each boundary point is connected to the centroid, producing a triangular fan that covers the interior region.  Adjacent triangles are paired to create quadrilaterals, which are then mapped to fixed-size square patches using affine warping. This procedure maintains both fine structural details and overall geometry. The obtained patches are then sequentially stitched together to form a merged representation, which is subsequently fed directly into the vision transformer as input tokens, enabling robust analysis of the ear’s structural features. 
\replythree{The warped patches could, in principle, directly be provided to a transformer as a token sequence. We instead reconstruct a $(112\times112)$ image so we can use a standard ViT without modification. In standard ViT implementations, tokens are produced internally from a fixed, non-overlapping grid (patch size = stride) and are tied to a fixed positional encoding. Feeding patches directly would therefore require an explicit token order and a custom positional scheme, which we avoid in this work.
}
A schematic illustration of the methodology is shown in Fig.~\ref{fig:M}.

The complete methodology, including preprocessing, contour extraction, point selection, triangular fan construction, warping to obtain square patches, and the creation of the merged representation, is described in detail below.

\begin{figure}[t]
\centering
  \includegraphics[width=0.9\linewidth,clip=]{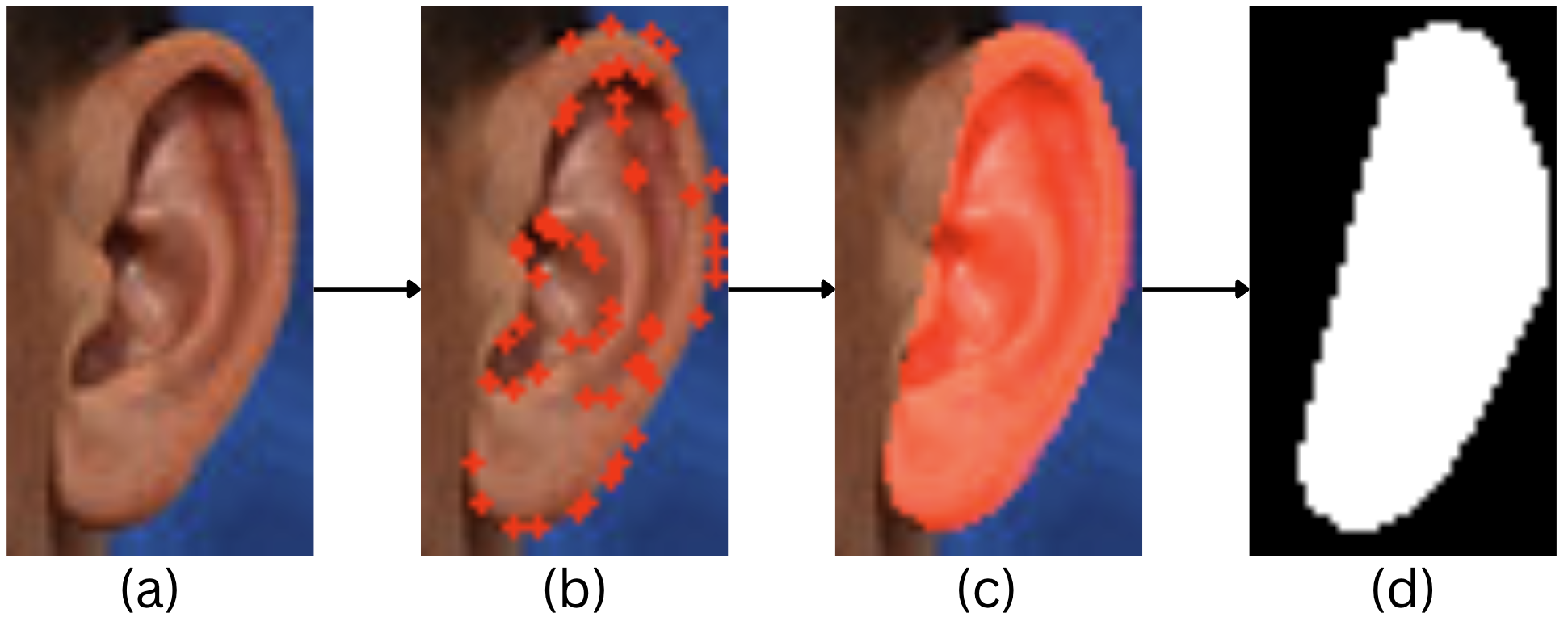}
  \caption{\textbf{Landmark map generation process:} (a) input ear image, (b) detected landmark points using 2SHG-Net model~\cite{hrovativc2023efficient}, (c) landmark map overlay on the ear image generated from the detected landmark points, and (d) final landmark map.}
  \label{fig:landmark_maps}
\end{figure}

\begin{figure}[t]
\centering
  \includegraphics[width=0.95\linewidth,clip=]{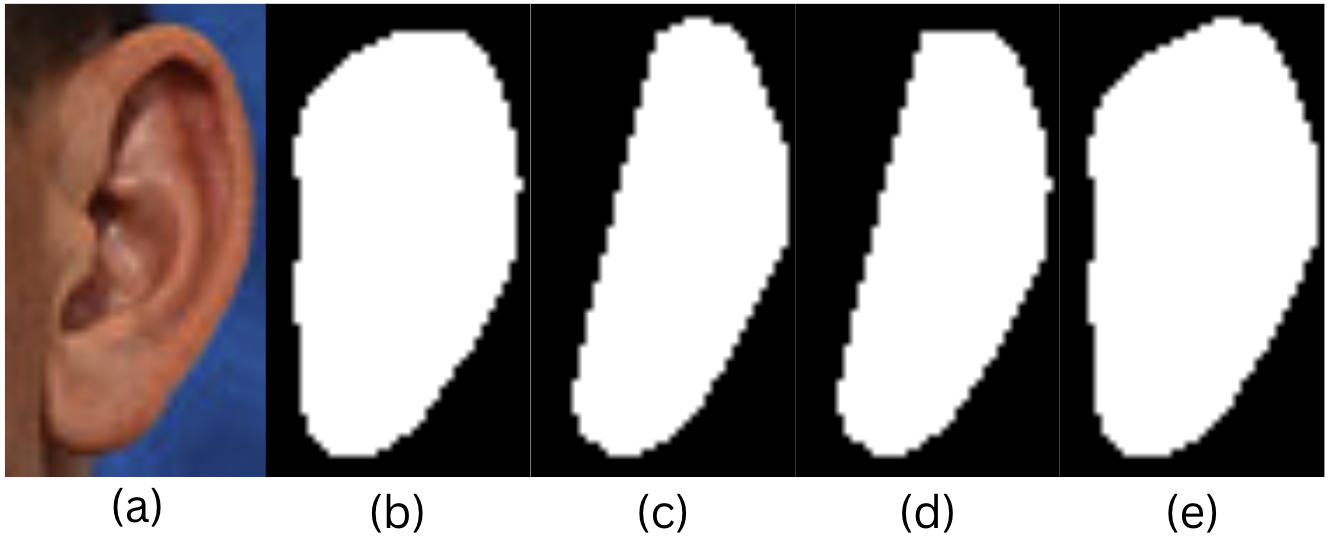}
  \caption{\textbf{Maps used in the patch-based warping process:} (a) original ear image, (b) segmentation map, (c) landmark map, (d) intersection map, and (e) union map obtained using the landmark and segmentation maps.}
  \label{fig:all_maps}
\end{figure}

\begin{figure*}[t]
  \centering
  \begin{subfigure}[t]{0.48\linewidth}
    \centering
    \includegraphics[width=\linewidth]{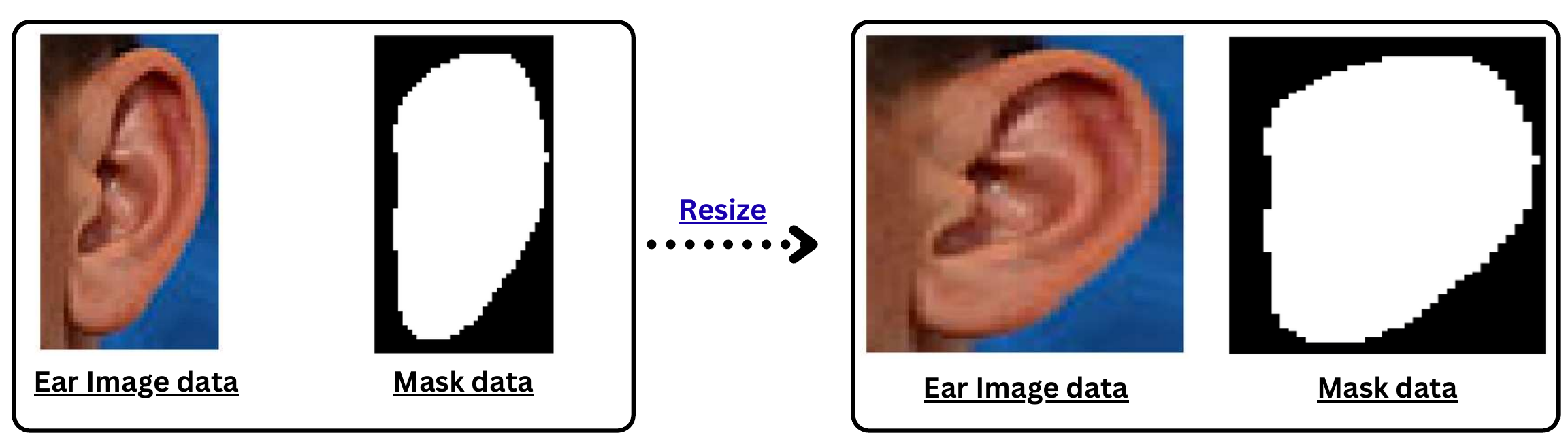}
    \caption{An example of the original and resized ear, along with its mask (segmentation mask).}
    \label{fig:M1}
  \end{subfigure}
  \hfill
  \begin{subfigure}[t]{0.48\linewidth}
    \centering
    \includegraphics[height=0.85in]{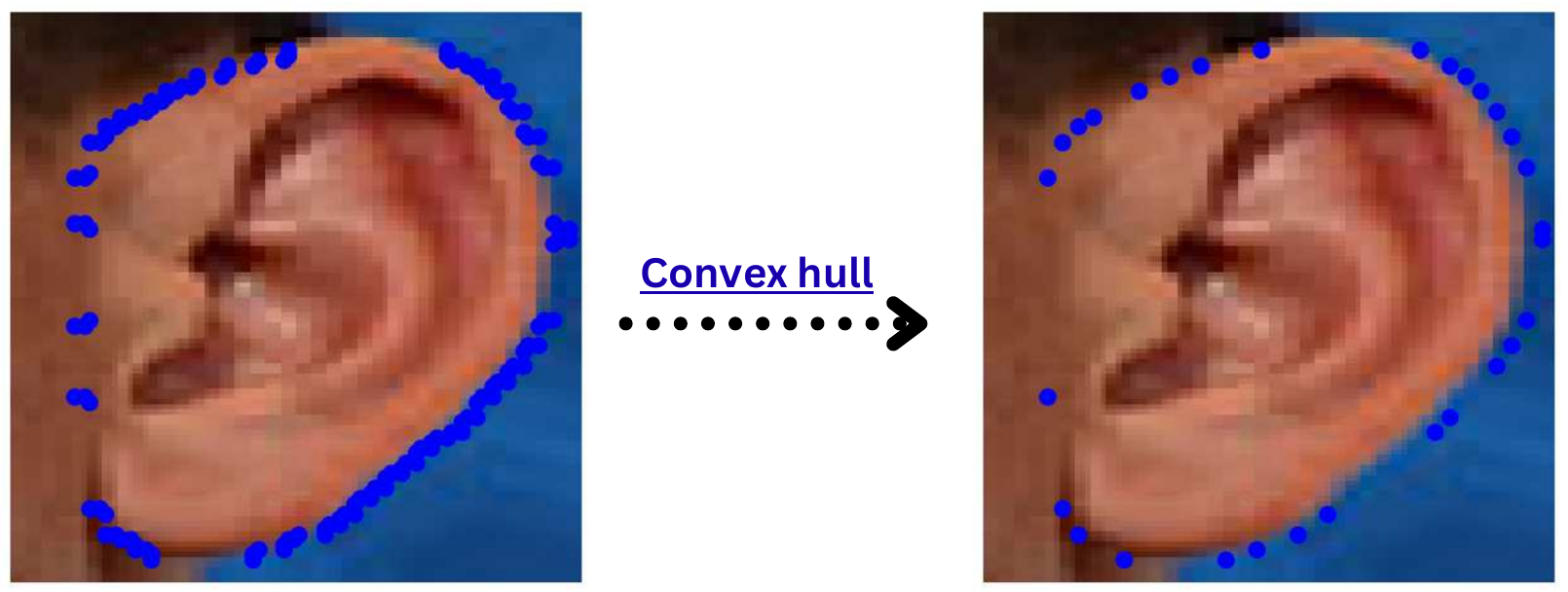}
    \caption{Find convex hull}
    \label{fig:M2}
  \end{subfigure}
  \hfill
  \begin{subfigure}[t]{0.48\linewidth}
    \centering
    \includegraphics[height=0.85in]{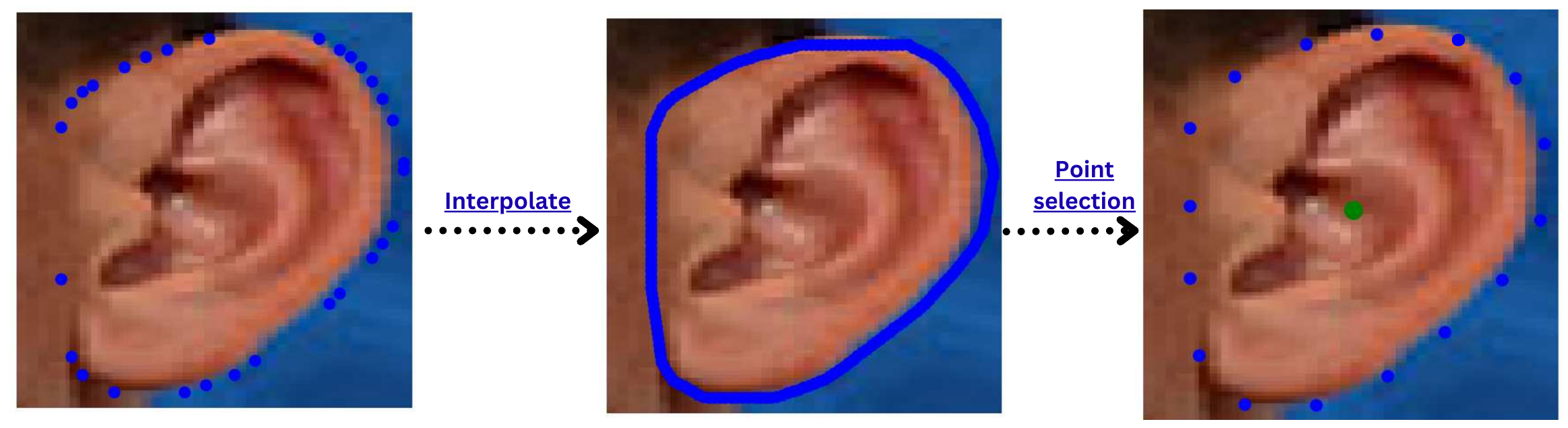}
    \caption{Extraction of centroid (green dot) and 16 equidistant points (blue dots)}
    \label{fig:M3}
  \end{subfigure}
  \hfill
  \begin{subfigure}[t]{0.48\linewidth}
    \centering
    \includegraphics[height=0.85in]{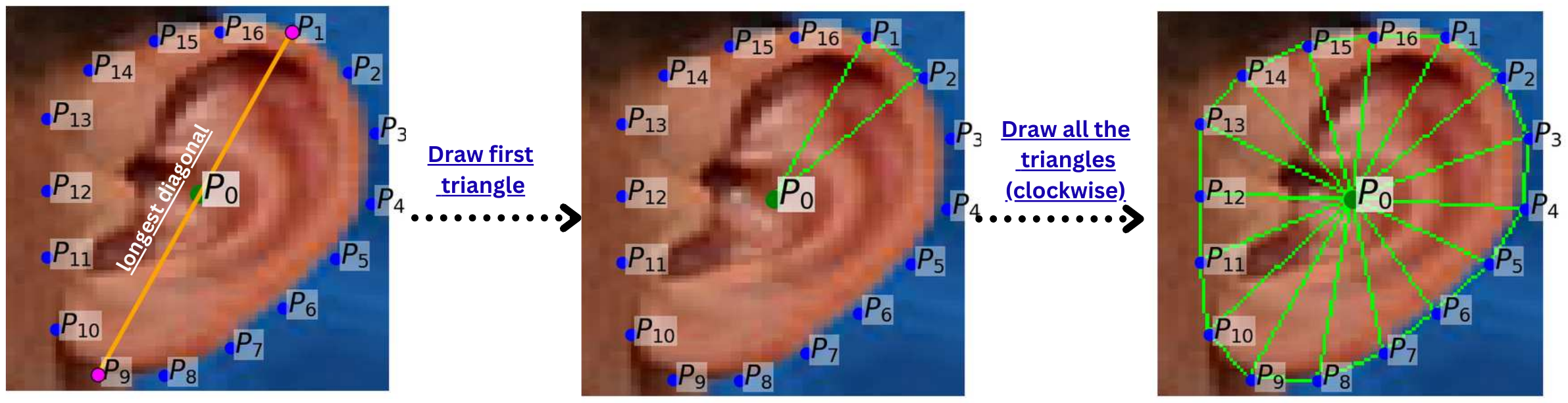}
    \caption{Triangular Fan Construction}
    \label{fig:M4}
  \end{subfigure}
  \hfill
  \begin{subfigure}[t]{0.36\linewidth}
    \centering
    \includegraphics[height=1.5in]{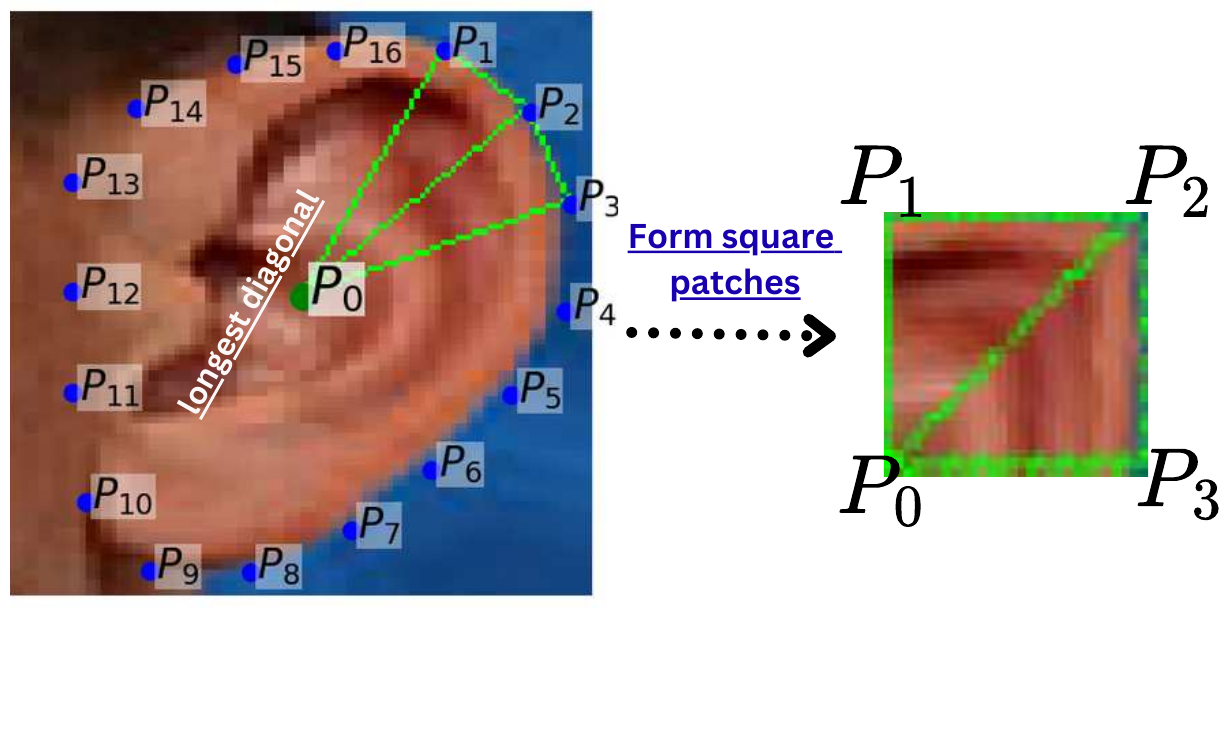}
    \caption{Quadrilateral Formation and Affine Warping}
    \label{fig:M5}
  \end{subfigure}
  \hfill
  \begin{subfigure}[t]{0.60\linewidth}
    \centering
    \includegraphics[height=2.4in]{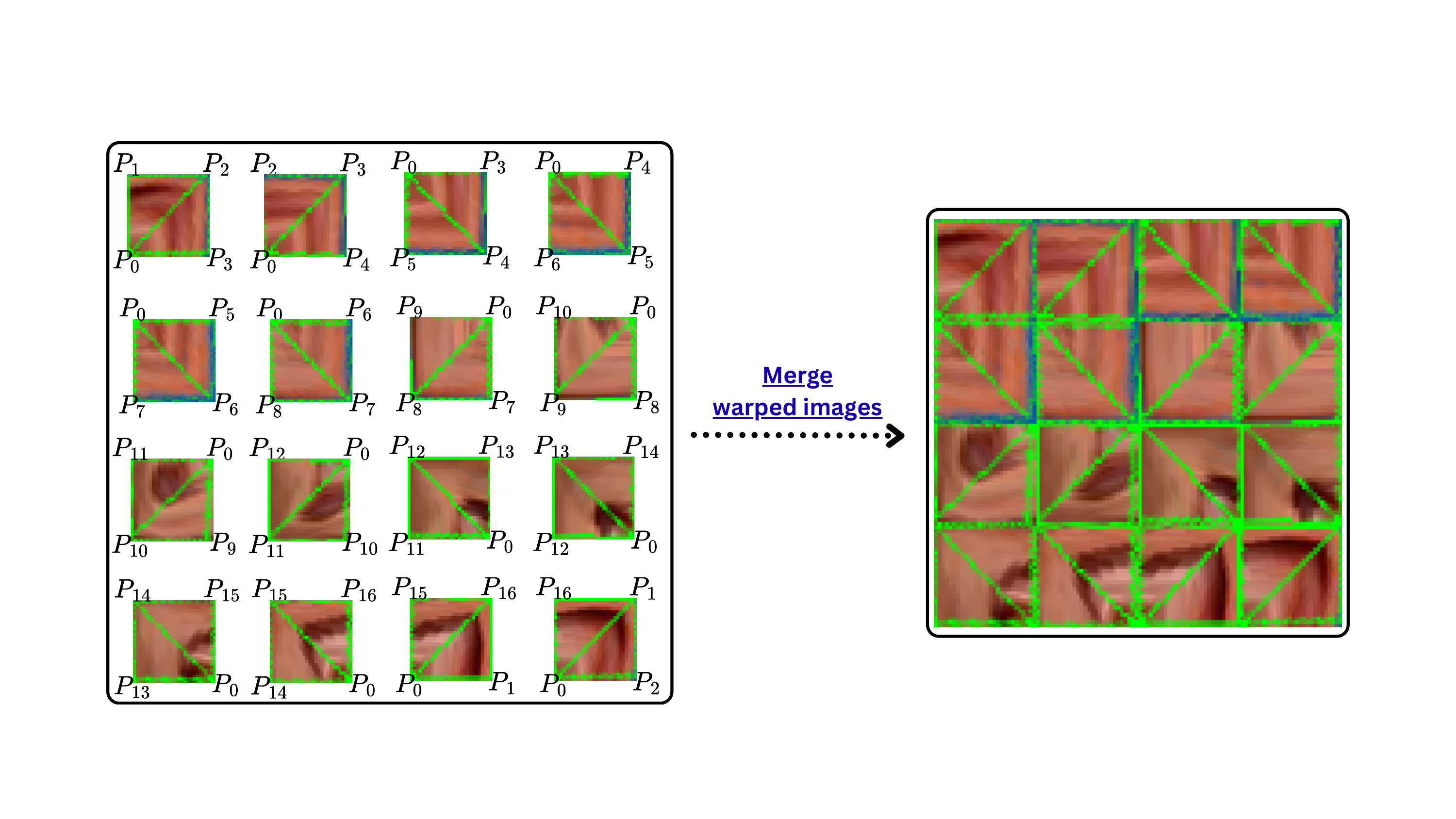}
    \caption{Patch Stitching}
    \label{fig:M6}
  \end{subfigure}
  \caption{Detailed methodology of the PaW-ViT algorithm. (\replythree{Note that the green lines and text-labels are added only to illustrate the flow between the steps, and they are not part of the method.})}
  \label{fig:M}
\end{figure*}

\subsection{Ear Image Preprocessing}
The ear images are segmented and cropped, and their corresponding binary masks are resized to a fixed dimension of $112 \times 112$ pixels. Fig.~\ref{fig:M1} shows an example of the original and resized ear along with its mask. We consider four types of masks as shown in Fig.~\ref{fig:all_maps} for four sets of experiments: (i) segmentation masks obtained using the BiSeNet architecture~\cite{bisenet}, trained on cropped ear images from CelebA-HQ~\cite{celebahq, celebahqmask} using the implementation from~\cite{bisenet_github}, (ii) landmark masks derived from the landmark points using the 2-SHGNet model~\cite{hrovativc2023efficient} as shown in Fig.~\ref{fig:landmark_maps}, (iii) intersection maps, and (iv) union maps. The latter two maps are computed from the segmentation and landmark masks. \replythree{Fig.~\ref{fig:all_maps}(b) shows the segmentation masks that are output directly by the BiSeNet model~\cite{bisenet} (i.e., it predicts a pixel-wise ear mask, not landmark points). In contrast, the landmark maps shown in Fig.~1 are built in two steps: the 2-SHGNet model~\cite{hrovativc2023efficient} first predicts landmark points and we then convert these points into a landmark mask for further processing.}

\subsection{Ear Boundary Extraction}
The boundary of the ear is initially extracted from the binary mask using contour detection. The boundary may have minor irregularities; therefore, to obtain a more reliable and refined approximation, the convex hull of the contour points is obtained. The convex hull provides a simplified and consistent outline of the ear boundary, serving as the basis for subsequent steps: optimized ear boundary sampling and triangular fan construction. The ear image with the convex hull overlay is shown in Fig.~\ref{fig:M2}.

\subsection{Ear Boundary Sampling and Centroid Computation}

To generate a dense set of boundary points, \replyone{the convex hull from the previous step is interpolated to 200 uniformly spaced samples. Before interpolation, we do not have a meaningful ``first" boundary point, since the ear contour is closed. We then fix a canonical reference on the contour as the topmost sample (minimum $y$; if tie then minimum $x$).}
From this optimized set of boundary points \replyone{and by uniform stepping from this reference,} 
16
equally spaced points
are then selected to ensure uniform coverage along the ear boundary. The centroid is computed using these 16
points. Incorporating the centroid will guarantee that each triangle accurately captures a segment of the ear’s interior structure. Fig.~\ref{fig:M3} shows the selected boundary points and the centroid on the ear image.

\subsection{Triangular Fan Construction}
\replyone{To establish a consistent reference frame for constructing the triangular fan, the selected boundary points are ordered in a clockwise sequence around the centroid. The ordering is determined by identifying the pair of boundary points with the largest Euclidean distance, referred to as the longest diagonal. From this pair, the point with the smaller vertical coordinate (closer to the top of the image) is chosen as the initial reference. The remaining points are then ordered clockwise around the centroid. This arrangement ensures reproducibility of the triangular fan and consistent orientation of adjacent triangles across different ear images.  
Using the centroid as the central base and the above clockwise direction in mind, triangles are formed by connecting it to each pair of successive boundary points. Specifically, for boundary points $\{P_1, P_2, \ldots, P_{16}\}$ and centroid $P_0$, the set of triangles is defined as $\{(P_0, P_i, P_{i+1}) : i = 1, \ldots, 16\},
$ where indexing is taken modulo $16$. The resulting triangles cover the entire ear interior, preserving structural details. Since we have 16 equidistant points, we get 16 triangles in total. Fig.~\ref{fig:M4} shows the triangular fan overlaying the sample ear image (Also showing the process for obtaining those 16 triangles).}

\subsection{Quadrilateral Formation and Affine Warping}
The unique clockwise ordering of the triangles enables
pairs of
adjacent triangles to form an ordered list
of
quadrilaterals, which will be individually warped into square patches using affine warping (the two adjacent triangles will share a (typically long) edge and can always be warped into a square). Affine transformations retain both collinearity and proportional distances, which drives the internal geometry of the triangles. The subsequent affine warps map each local sector into a fixed-resolution warped patch of $28 \times 28$ pixels for uniformity of token size.
In total, we obtain 16 patches. For each warped patch, the triangle vertices are reordered or rotated so that the orientation of the common edge matches the orientation in the warped-square patch, thereby retaining the ear’s internal structure.
\replythree{Note that the affine transformation involves resampling, so some smoothing can occur, especially when the mapping is strong or when the output is low-resolution. In our setup, each region is a triangular sector with a regular, half-square like geometry, so the resulting deformation is generally limited and is not expected to remove substantial fine detail.}
Fig.~\ref{fig:M5} shows an example of two triangles being warped and combined into a square patch.

\subsection{Patch Stitching}
The 16 warped square patches, each resized to $28\times28$ pixels, are sequentially arranged into a  $4\times4$ grid to reconstruct a unique $112\times 112$ canvas. The ordering follows a row-wise sequence: the first four patches fill the top row, the next four fill the second row, and so forth.
The obtained canvas provides a reproducible, anatomy-aware representation that captures both local triangular features and the overall ear structure. An example of the completed tiled canvas is presented in Fig.~\ref{fig:M6}.
Unlike the standard ViT approach, where patches are taken directly from the raw images, our method ensures anatomical alignment before processing. This reduces inconsistencies caused by variations in ear shape, size, or pose. 
As a result, the input features fed into the ViT remain more stable across samples, leading to stronger and more reliable embeddings for verification. \replythree{The green lines in Fig.~\ref{fig:M6} are added only to illustrate the flow between the steps, and they are not part of the method and are not used in any computation.}

\section{Datasets}
\label{sec:datasets}

\subsection{Training data}

The \textbf{UERC2023} dataset~\cite{emervsic2023unconstrained}
\footnote{Available at \url{[http://uerc.fri.uni-lj.si/uerc.html}} was used to train the vision transformers. It offers a large-scale, diverse, and unconstrained collection of ear images, making it well-suited for robust model training. It integrates UERC2017 and UERC2019 (14,004 images, 650 subjects) with VGGFace-Ear~\cite{ramos2022vggface} (234,651 images, 660 subjects), generated by cropping ear regions from VGGFace~\cite{cao2018vggface2}. To prevent overlap with the AWE~\cite{emervsivc2017ear} test set, the first 100 subjects were excluded. 

\subsection{Testing data}

Evaluation was conducted on four independent datasets: OPIB, AWE, WPUT, and EarVN1.0.

\begin{enumerate}

\item \textbf{OPIB} ~\cite{Adebayo2023}: Consists of 907 images of 152 African subjects, left and right ears captured at 0°, 30°, and 60°. Includes occlusions (headphones, earrings, scarves) and ear images taken indoors and outdoors. Gender distribution: 59.65\% male, 40.35\% female.

\item \textbf{AWE}~\cite{emervsivc2017ear}: Comprises 1,000 web-collected images of 100 celebrities (10 per subject). Annotations include gender, ethnicity, accessories, occlusion levels (65\% none, 28\% mild, 7\% severe), head pose, and ear side.

\item \textbf{WPUT}~\cite{frejlichowski2010west}: Includes 2,071 images from 501 subjects with a balanced gender distribution. Each subject has at least four images captured at 90° and 75° poses. Around 80\% of images feature occlusions (hair, earrings, glasses), and 8\% exhibit motion blur.

\item \textbf{EarVN1.0}~\cite{hoang2019earvn1}: Consists of 28,412 low-resolution images from 164 Asian subjects (over 100 images per subject), captured under varying illumination, occlusion, and pose conditions. As the largest and most challenging dataset due to significant variations in pose and resolution, it is used solely for testing to facilitate cross-dataset evaluation.

\end{enumerate}

\section{Experimental Setup}
\label{sec:exp}

The ear recognition models were trained in a verification scenario using four different configurations of Vision Transformers (ViTs)--ViT-T (Tiny), ViT-S (Small), ViT-B (Base), and ViT-L (Large)--on the UERC2023 dataset (excluding the first 100 subjects that belong to the AWE dataset). The dataset was initially pre-processed using a ResNet-100 based ear-side classifier to distinguish between the left and right ear, treating each ear side as a separate entity. As a result, each subject had two separate folders: one for their right ear and another for their left ear, with each side treated as an independent subject.

In addition to the baseline experiments using the original, unprocessed images, four additional sets of experiments were conducted using images processed with segmentation maps, landmark maps, intersection maps, and union maps. These were designed to evaluate which representation provides the optimal ear region for recognition. As described in Sec.~\ref{sec:methodology}, each ear image undergoes an anatomy-aware preprocessing using our patch-based warping method that converts the ear geometry into a structured patch representation.
This anatomically informed representation is subsequently fed into the Vision Transformers, where each patch is linearly projected and processed by the encoder layers. \replycommon{To strengthen the comparison, we also report a simple yet competitive baseline based on background masking with standard rectangular patches (Table~\ref{tab:model_metrics}, \textit{Background masking}), where all pixels outside the segmentation mask are set to black.}

\replycommon{ All Vision Transformer variants are trained for 100 epochs using the same optimization protocol to ensure fair comparison across model scales. Input images are resized to \(112\times112\), and each model employs standard patch-based tokenization with a transformer encoder of variant-specific depth and embedding dimension. Regardless of the internal embedding size, all models project features to a shared 512-dimensional embedding space for supervision. Training is conducted from scratch using the AdamW optimizer with a learning rate of \(1\times10^{-3}\) and a weight decay of 0.1, with a fixed batch size of 128 and mixed-precision (FP16) enabled for computational efficiency. A sample rate of 0.3 is used, and no pretrained checkpoints are employed. Supervision is provided via a margin-based loss with margins \((1.0, 0.0, 0.4)\), while regularization is applied through stochastic depth with variant-specific drop-path rates, and token masking is disabled during training.} The models were trained for each ViT configuration considering a patch size of $28\times28$. All models used non-overlapping patches, with strides equal to the patch size. To assess the performance of the trained models, ear matching experiments were conducted on four separate test datasets: OPIB, AWE, WPUT, and EarVN1.0. For each experiment, the entire dataset was used for testing, which differs from prior research, where datasets were split into training and testing sets.

Ear matching was performed by computing the dot product similarity scores of the 512-dimensional feature embeddings extracted from the ViT models. We construct evaluation pairs following standard biometric protocols: for dataset with $N$ subjects and $M_i$ images per subject $i$ (for $i=1,2,3,...,N$), we generate $\sum_{i=1}^{N} \binom{M_i}{2}$ genuine pairs and $\sum_{i=1}^{N-1}\sum_{j=i+1}^N M_i M_j$ impostor pairs. The model’s performance was assessed using the AUC score, which reflects its ability to differentiate between genuine and impostor pairs by analyzing the ROC curve, showing the trade-off between True Positive Rate (TPR) and False Positive Rate (FPR). The matching experiment was repeated five times to ensure the reliability of the results. The average Area Under the Curve (AUC) scores, along with empirical confidence intervals for each configuration, were computed and are presented in Table~\ref{tab:model_metrics}.

\section{Results}
\label{sec:results}

\renewcommand{\arraystretch}{1.4}
\newcommand{\near}{\cellcolor{yellow!20}}   
\newcommand{\beat}{\cellcolor{green!20}}    

\begin{table*}[t]
  \centering
  \small
  \caption{\MakeUppercase{Comparison of AUC scores across map types (Original, landmark, segmentation, union, intersection) for ViT\_T, ViT\_S, ViT\_B, and ViT\_L on OPIB, AWE, WPUT, and EarVN1.0. Cells in yellow indicate the closest score to the baseline without surpassing it; cells in green indicate scores that surpass the baseline.}}
  \label{tab:model_metrics}
  \begin{tabular}{l lcccc}
    \toprule
    \textbf{Model} & \textbf{Map Type} & \textbf{OPIB} & \textbf{AWE} & \textbf{WPUT} & \textbf{EarVN1.0} \\
    \midrule
    \multicolumn{6}{l}{\textit{Tiny (ViT\_T)}} \\
    \multirow{6}{*}{ViT\_T} & None (Baseline)         & 0.8926 $\pm$ 0.0055 & 0.9732 $\pm$ 0.0015 & 0.9286 $\pm$ 0.0047 & 0.7356 $\pm$ 0.0017 \\
    & \replytwo{Background masking}    & 0.8770 $\pm$ 0.0043 & 0.9674 $\pm$ 0.0024 & 0.8830 $\pm$ 0.0040 & 0.7160 $\pm$ 0.0020 \\
                            & Segmentation maps & \near 0.8770 $\pm$ 0.0100 & \near 0.9696 $\pm$ 0.0011 & \near 0.8960 $\pm$ 0.0050 & \near 0.7202 $\pm$ 0.0022 \\
                            & Landmark maps     & 0.8260 $\pm$ 0.0050 & 0.9382 $\pm$ 0.0035 & 0.8650 $\pm$ 0.0038 & \beat 0.7460 $\pm$ 0.0030 \\
                            & Union maps        & 0.8430 $\pm$ 0.0082 & \beat 0.9760 $\pm$ 0.0020 & \near 0.8960 $\pm$ 0.0030 & \beat 0.7820 $\pm$ 0.0020 \\
                            & Intersection maps & 0.8218 $\pm$ 0.0046 & 0.9390 $\pm$ 0.0030 & 0.8658 $\pm$ 0.0028 & \beat 0.7640 $\pm$ 0.0020 \\
    \midrule
    \multicolumn{6}{l}{\textit{Small (ViT\_S)}} \\
    \multirow{5}{*}{ViT\_S} & None (Baseline)          & 0.8934 $\pm$ 0.0076 & 0.9640 $\pm$ 0.0019 & 0.9270 $\pm$ 0.0017 & 0.7164 $\pm$ 0.0017 \\
    & \replytwo{Background masking}    & 0.8820 $\pm$ 0.0080 & 0.9602 $\pm$ 0.0013 & 0.8840 $\pm$ 0.0080 & 0.7030 $\pm$ 0.0033 \\
                            & Segmentation maps & \near 0.8660 $\pm$ 0.0090 & \near 0.9570 $\pm$ 0.0040 & \near 0.8830 $\pm$ 0.0032 & 0.7000 $\pm$ 0.0050 \\
                            & Landmark maps     & 0.8170 $\pm$ 0.0080 & 0.9152 $\pm$ 0.0070 & 0.8510 $\pm$ 0.0080 & \beat 0.7210 $\pm$ 0.0050 \\
                            & Union maps        & 0.8170 $\pm$ 0.0040 & 0.9162 $\pm$ 0.0042 & 0.8630 $\pm$ 0.0040 & \near 0.7050 $\pm$ 0.0050 \\
                            & Intersection maps & 0.8238 $\pm$ 0.0103 & 0.9188 $\pm$ 0.0038 & 0.8616 $\pm$ 0.0108 & \beat 0.7380 $\pm$ 0.0030 \\
    \midrule
    \multicolumn{6}{l}{\textit{Base (ViT\_B)}} \\
    \multirow{5}{*}{ViT\_B} & None (Baseline)          & 0.9036 $\pm$ 0.0040 & 0.9044 $\pm$ 0.0032 & 0.9320 $\pm$ 0.0019 & 0.7278 $\pm$ 0.0033 \\
    & \replytwo{Background masking}    & 0.8828 $\pm$ 0.0040 & 0.9512 $\pm$ 0.0016 & 0.8840 $\pm$ 0.0050 & 0.7010 $\pm$ 0.0030 \\
                            & Segmentation maps & \near 0.8978 $\pm$ 0.0095 & \beat 0.9720 $\pm$ 0.0020 & \near 0.9070 $\pm$ 0.0040 & \near 0.7240 $\pm$ 0.0150 \\
                            & Landmark maps     & 0.8550 $\pm$ 0.0040 & \beat 0.9380 $\pm$ 0.0030 & 0.8736 $\pm$ 0.0051 & \beat 0.7398 $\pm$ 0.0024 \\
                            & Union maps        & 0.8770 $\pm$ 0.0050 & \beat 0.9750 $\pm$ 0.0020 & 0.9042 $\pm$ 0.0060 & \beat 0.7620 $\pm$ 0.0030 \\
                            & Intersection maps & 0.8572 $\pm$ 0.0079 & \beat 0.9392 $\pm$ 0.0037 & 0.8810 $\pm$ 0.0040 & \beat 0.7530 $\pm$ 0.0040 \\
    \midrule
    \multicolumn{6}{l}{\textit{Large (ViT\_L)}} \\
    \multirow{5}{*}{ViT\_L} & None (Baseline)         & 0.9028 $\pm$ 0.0042 & 0.9610 $\pm$ 0.0037 & 0.9272 $\pm$ 0.0048 & 0.7176 $\pm$ 0.0046 \\
    & \replytwo{Background masking}    & 0.8800 $\pm$ 0.0050 & 0.9498 $\pm$ 0.0027 & 0.8820 $\pm$ 0.0050 & 0.6960 $\pm$ 0.0030 \\
                            & Segmentation maps & \near 0.8960 $\pm$ 0.0060 & \beat 0.9690 $\pm$ 0.0030 & \near 0.9040 $\pm$ 0.0071 & \near 0.7090 $\pm$ 0.0050 \\
                            & Landmark maps     & 0.8558 $\pm$ 0.0045 & \near 0.9310 $\pm$ 0.0040 & 0.8756 $\pm$ 0.0068 & \beat 0.7322 $\pm$ 0.0030 \\
                            & Union maps        & 0.8830 $\pm$ 0.0080 & \beat 0.9760 $\pm$ 0.0040 & 0.9000 $\pm$ 0.0080 & \beat 0.7600 $\pm$ 0.0030 \\
                            & Intersection maps & 0.8590 $\pm$ 0.0110 & 0.9302 $\pm$ 0.0040 & 0.8730 $\pm$ 0.0020 & \beat 0.7450 $\pm$ 0.0020 \\
    \bottomrule
  \end{tabular}
\end{table*}


\subsection{Model-based Performance Results}

In this study, we evaluated the performance of different Vision Transformer (ViT) models, including ViT-Tiny (ViT-T), ViT-Small (ViT-S), ViT-Base (ViT-B), and ViT-Large (ViT-L), across four benchmark datasets: OPIB, AWE, WPUT, and EarVN1.0. Table~\ref{tab:model_metrics} summarizes the AUC scores obtained by each model under the baseline configuration. The results show that the ViT-B and ViT-L models generally outperformed the smaller variants, achieving more stable and competitive results across most datasets (see Fig.~\ref{fig:max_auc_scores}). Specifically, ViT-B achieved the highest AUC on OPIB (0.9036 $\pm$ 0.0040) and WPUT (0.9320 $\pm$ 0.0019), while ViT-T slightly outperformed others on AWE with 0.9760 $\pm$ 0.0020 and EarVN1.0 with 0.7820 $\pm$ 0.0020. 

Among the ViT-L models, performance remained competitive across datasets, with strong results on AWE (0.9760 $\pm$ 0.0040) and WPUT (0.9272 $\pm$ 0.0048). ViT-S models, while showing consistent performance, did not surpass the larger configurations but maintained competitive AUC values such as 0.9640 $\pm$ 0.0019 on AWE and 0.9270 $\pm$ 0.0017 on WPUT. By contrast, ViT-B, despite being a mid-sized model, achieved robust performance on OPIB and WPUT but fell slightly behind on AWE compared to ViT-L and ViT-T.

In conclusion, ViT-B and ViT-L demonstrated the strongest overall performance across datasets, ViT-T showed competitive results with advantages on AWE and EarVN1.0, and ViT-S remained stable but less dominant compared to the other configurations. \replythree{ Since we train on UERC2023 and evaluate cross-dataset on four disjoint test sets, model capacity does not guarantee dominance under distribution shift; smaller models can sometimes generalize better on certain test distributions while larger models can be more sensitive to mismatch. Importantly, larger models remain strong overall baselines, and ViT-B achieves the best baseline performance on OPIB and WPUT and ViT-L on WPUT, suggesting that the observed variability is more consistent with cross-dataset generalization effects.}

\subsection{Dataset-based Performance Results}

\begin{figure*}[t]
\centering
  \includegraphics[width=0.95\linewidth,clip=]{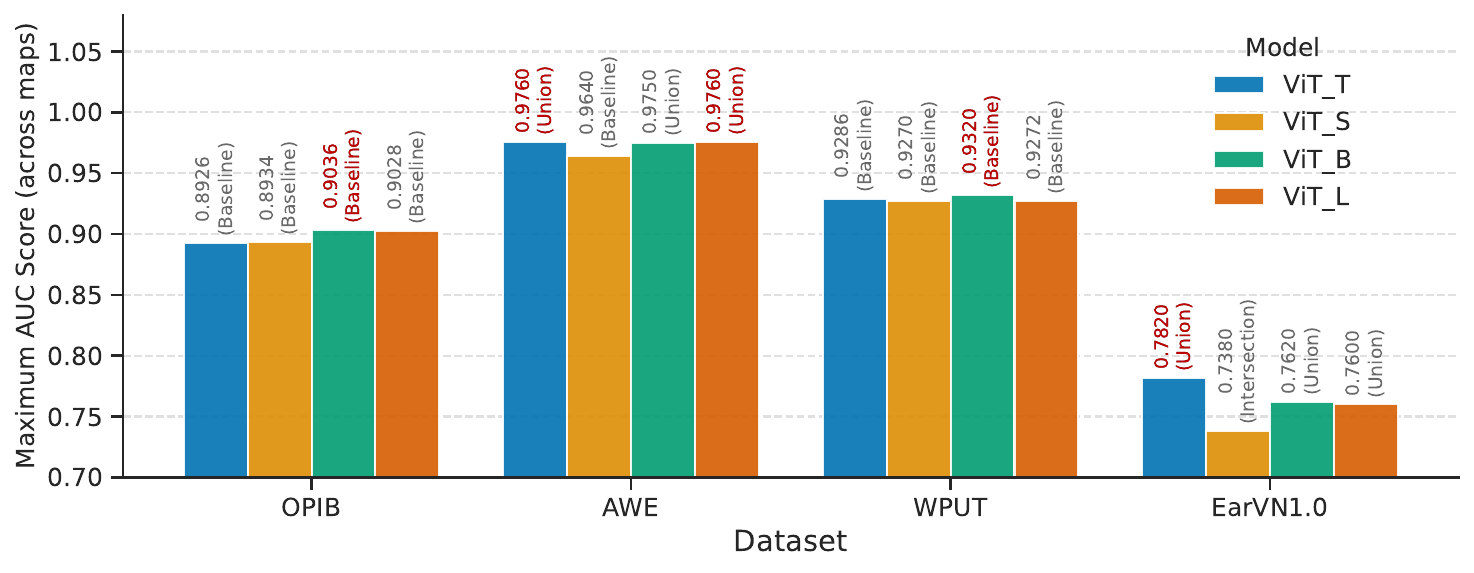}
  \caption{{Peak performance of ViT models across datasets. Each bar's label indicates the maximum AUC score and the specific map type that achieved it. The AUC score (and its label) for the top-performing model for each dataset is highlighted in red.}}
  \label{fig:max_auc_scores}
\end{figure*}

Across datasets, AWE consistently achieved the highest AUC values for all models, with baselines exceeding 0.96. OPIB and WPUT also demonstrated strong baseline performance; however, none of the processed representations were able to surpass the baseline scores on these two datasets (see Fig.~\ref{fig:max_auc_scores}). This limitation may stem from the presence of numerous ear accessories, which introduce occlusions and distortions that reduce the effectiveness. In contrast, EarVN1.0 remained the most challenging dataset overall, with baseline AUC values ranging from 0.7380 (ViT\_S) to 0.7820 (ViT\_T). Notably, several patch-based warping processed representations, particularly using union and intersection maps, surpassed the baseline on EarVN1.0. \replycommon{These results clarify when anatomically-aware preprocessing is beneficial. On EarVN1.0, strong unconstrained variation (pose/illumination/resolution) makes canonicalization reduce geometric mismatch and better support patch tokenization and attention. In contrast, OPIB/WPUT include frequent accessory-driven occlusions that distort anatomical boundaries, reducing the reliability of boundary-driven warps and leaving little room for consistent gains; accordingly, union/intersection warping is most helpful on EarVN1.0 but remains near (or slightly below) baseline on OPIB/WPUT.}

\subsection{Map-based Warped Image Performance Results}

The effect of different map-based warped images varied substantially across models and datasets. \replytwo{For a broader comparison, we include a stronger, simpler baseline based on background masking with standard rectangular patches (Table~\ref{tab:model_metrics}, \textit{Background masking}), confirming that the observed gains are not attributable to naïve masking alone.}
Segmentation map–based warped images consistently approached baseline performance but rarely surpassed it, indicating their limited yet stable contribution. Landmark map–based warped images generally underperformed relative to the baseline, though they occasionally exceeded it on EarVN1.0. Union map–based warped images delivered the most notable improvements, frequently surpassing baseline performance, especially on AWE and EarVN1.0. Intersection map–based warped images also showed competitive performance, often outperforming the baseline on EarVN1.0 
and achieving comparable results on other datasets. However, for OPIB and WPUT, none of the map-based warped images surpassed baseline performance (Fig.~\ref{fig:max_auc_scores}), further confirming that the heavy presence of ear accessories in these datasets constrains the benefits of patch-based warping. Collectively, these results suggest that while segmentation- and landmark-based warped images alone provide limited gains, combining multiple map-based warped images (union or intersection) yields complementary information that surpasses baseline performance -- particularly in datasets with fewer structured obstructions such as AWE and EarVN1.0.

\section{Conclusion}
\label{sec:conclusion}

In this paper, we introduced \textbf{PaW-ViT}, a patch-based warping vision transformer that uses anatomy-aware preprocessing for robust ear recognition. The proposed framework addresses two key challenges in applying Vision Transformers to biometric tasks: (i) the misalignment of anatomical structures when raw images are directly tokenized, and (ii) the tendency of models to attend to irrelevant background regions. By sampling ear boundaries, constructing triangular fans, performing affine warping, and stitching patches into structured grids, PaW-ViT ensures anatomically consistent, background-free representations that preserve both fine-grained details and global structural continuity. \replyone{To reduce sensitivity to local boundary noise, we refined the raw contour using the convex hull and then interpolated it densely (200 samples) before selecting 16 equidistant anchors, which stabilized point placement and yielded reproducible fan partitions across samples. The subsequent affine warps map each local sector into a fixed-resolution patch under a deterministic ordering, ensuring the resampling is applied consistently across images.} This enables more stable embeddings and enhances the discriminative capacity of ViTs for ear biometrics. 

Comprehensive experiments across four benchmark datasets--OPIB, AWE, WPUT, and EarVN1.0--yield several important observations. First, \textbf{model-level performance analysis} shows that ViT-B and ViT-L consistently provide the strongest overall baselines, with ViT-B excelling on OPIB and WPUT, while ViT-L achieves competitive stability across datasets. Interestingly, ViT-T, despite its smaller size, surpassed larger variants on AWE and EarVN1.0, demonstrating that smaller models can remain highly competitive when paired with anatomically informed preprocessing. ViT-S, while stable, did not scale as effectively.  

Second, \textbf{dataset-level analysis} highlights the role of dataset characteristics. AWE achieved the highest baseline scores (AUC $>$ 0.96) across all models which improved to 0.9760 using PaW-ViT approach. In contrast, OPIB and WPUT, heavily influenced by ear accessories and occlusions, limited the benefits of warping. EarVN1.0 emerged as the most challenging dataset, with relatively low baselines (AUC $\sim$0.71--0.73). On this dataset, PaW-ViT delivered the most significant improvements, with union and intersection map based warped images consistently surpassing baseline scores. These results confirm that map-based warping is particularly effective in less structured or more variable scenarios.  

Third, \textbf{map-based warping performance} revealed distinct contributions. Segmentation map based warped images reliably approached baseline performance, suggesting stable yet modest improvements. Landmark map based warped images underperformed overall, though they occasionally surpassed baselines in EarVN1.0. \replycommon{
We also evaluated their union and intersection and observed that the combined maps are more robust than single-source guidance when the two maps fail differently (e.g., segmentation can miss thin structures, while landmarks can be sparse or slightly shifted). Consistently, the union and intersection variants are stronger than landmark-only and often recover performance when segmentation-only performance degrades, indicating that complementary cues mitigate individual failure modes.
} Union map based warped images produced the most consistent gains, particularly on AWE and EarVN1.0, while intersection map based warped images also performed strongly, frequently surpassing baselines on EarVN1.0.  

In summary, PaW-ViT establishes a novel anatomy-aware preprocessing framework for ear recognition, significantly improving performance in cases where baseline ViTs struggle with variability and occlusion. Larger ViT architectures deliver robust baselines, but map-based warping provides consistent advantages, especially in challenging datasets. \textbf{Future work} will also explore adaptive boundary sampling and robustness to occlusions to further enhance generalization in unconstrained scenarios. This study demonstrates that geometry-driven preprocessing, when integrated with transformer architectures, represents a promising pathway toward reliable, anatomy-aware biometric recognition systems.






{\small
\bibliographystyle{ieee}
\bibliography{egbib}
}

\end{document}